%% file: main.tex
\documentclass[letterpaper, 10 pt, conference]{ieeeconf}
\IEEEoverridecommandlockouts      
\usepackage[dvipsnames,table,xcdraw]{xcolor}
\usepackage[ruled,vlined]{algorithm2e}
\usepackage{graphics} 
\usepackage{graphicx} %
\usepackage{lipsum,adjustbox}
\usepackage{tikz}
\usetikzlibrary{positioning, shapes.geometric}
\usepackage{verbatim}
\usetikzlibrary{calc}

\usepackage[caption=false,font=footnotesize]{subfig}
\usepackage{float}
\usepackage{amsmath} 
\usepackage{amssymb}  
\usepackage{mathtools}

\usepackage{mymath,gmech}
\usepackage{cite}
\let\labelindent\relax
\usepackage{enumitem}
\usepackage{mleftright}

\usepackage{multirow}
\graphicspath{ {../figs/}{../figures/}}

\newcommand{\MRSEdit}[1]{\textcolor{black}{#1}}

\newcommand{\toedit}[1]{\textcolor{black}{#1}}

\newcommand{\Manishawrite}[1]{\textcolor{green}{#1}}

\newcommand{\ST}{S\&T}

\makeatletter 
  \newcommand\figcaption{\def\@captype{figure}\caption} 
  \newcommand\tabcaption{\def\@captype{table}\caption} 
\makeatother

\title{\LARGE \bf Adversarial Search and Tracking with Multiagent Reinforcement Learning in Sparsely Observable Environment}

 \author{{Zixuan Wu$^{1,\dagger}$, Sean Ye$^{1,\dagger}$, Manisha Natarajan$^{1}$, Letian Chen$^{1}$, Rohan Paleja$^{1}$, and Matthew C. Gombolay$^{1}$}
 \thanks{*This work is supported in part by the Office of Naval Research (ONR) under grant numbers N00014-19-1-2076, N00014-22-1-2834, and N00173-21-1-G009, the National Science Foundation under grant CNS-2219755, and MIT Lincoln Laboratory grant number 7000437192.}
 \thanks{$\dagger$ Equal contribution}
 \thanks{$^{1}$All authors are associated with the Institute of Robotics and Intelligent Machines (IRIM), Georgia Institute of Technology, Atlanta, GA 30308, USA.}
\thanks{Correspondance Author: Zixuan Wu 
{\tt\small zwu380@gatech.edu}}
 }
\begin{document}

\maketitle
\thispagestyle{empty}
\pagestyle{empty}

\begin{abstract}

We study a search and tracking (\ST) problem where a team of dynamic search agents must collaborate to track an adversarial, evasive agent. The heterogeneous search team may only have access to a limited number of past adversary trajectories within a large search space. This problem is challenging for both model-based searching and reinforcement learning (RL) methods since the adversary exhibits reactionary and deceptive evasive behaviors in a large space leading to sparse detections for the search agents. To address this challenge, we propose a novel Multi-Agent RL (MARL) framework that leverages the estimated adversary location from our learnable filtering model. We show that our MARL architecture can outperform all baselines and achieves a 46\% increase in detection rate. 

\color{black}

\end{abstract}


\input{intro.tex}

\input{lit_review.tex}

\input{background.tex}

\input{method.tex}

\input{result.tex}

\input{conc.tex}

\bibliographystyle{IEEEtran}
\bibliography{targetSearch, filter,MARL}



\end{document}

%% file: intro.tex
\section{Introduction \label{sec:Intro}}

Search and tracking (\ST) is a fundamental problem that requires autonomous agents to work together to find and track a specific object of interest, which is critical for a wide range of applications, from drug smuggling tracking to endangered wildlife monitoring. It is often difficult and cumbersome to design hand-crafted expert \ST{} policies as it is intractable to manually abstract patterns from complex environments with limited knowledge of adversarial target motions perfectly. Therefore, we propose a statistic-driven MARL framework with a Prior Motion Combined (PMC) filter model to perform informed exploration and maximize performance in a complex sparsely observable environment.

 In this paper, we focus on the challenging adversarial \ST{} setting of reactive target and sparse detection, which means our heterogeneous search team can only make decisions based on intermittent information. Previous non-learning based methods have been developed to provide fast, robust and accurate tracking capabilities \cite{activeSearch, kwa2020optimal, kennedy1995particle, eberhart2001tracking, liu2016topology}, which sometimes requires auxiliary parallel and hierarchical architectures, including probability map construction \cite{strategies}, detection-object association \cite{IMM0, IMM1, multiaMultio}, target location filtering \cite{IMM0, IMM1, taxi}, and searching strategy optimization \cite{multiAUV, strategies}. The appropriate design, organization, and coordination of such subsystems can improve the \ST{} efficiency, but none of these is designed for {\textbf{adversarial}} \ST{}. In recent years, MARL has shown promising results in learning \ST{} policies by outperforming hand-tuned expert heuristics \cite{multiAUV}. Several approaches have attempted to tackle exploration in sparse reward environments through reward shaping \cite{xie2019deep} or experience replay mechanisms \cite{du2022lucid, zhou2023cooperative}. However, these methods still rely on expert-designed rewards, goals, or repeated environment replay from specific states. In contrast, we propose to address this challenge using opponent modeling, which involves reasoning about the opponent's strategy to enhance decision-making \cite{von2017minds}.

In this work, we present a novel approach that leverages estimated adversarial locations from the PMC filter to improve the efficiency of MARL for \ST{} in large adversarial environments with sparse \MRSEdit{detections}. The PMC filtering model consists of a mixture-density network (MDN) that balances prior knowledge with a forward-motion model and is trained from past information we have about the target (e.g. wild animal behavior patterns \cite{reader2003social} or drug smuggling routine \cite{medel2015mexico} etc.) The key idea is to enhance the RL process by introducing a filtering model, which serves as a shortcut for inferring the evading agent's location. 



\begin{figure*}[t]
  \vspace*{0.06in}
  \centering

  \includegraphics[width=0.9\textwidth]{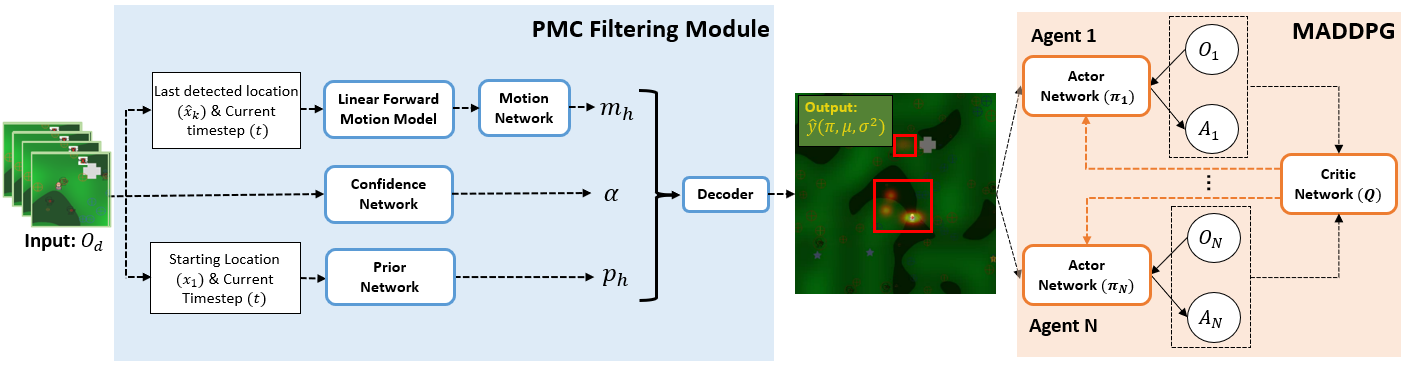}
\vspace{-0.25cm}
  \caption{\toedit{Proposed Architecture (PMC-MADDPG): We augment the observations of each search agent with outputs from our Prior-Motion Combined (PMC) filtering model - gold regions (e.g. in the red boxes) indicate multiple output Gaussians on the map. The PMC is warmstarted offline and trained using the negative log likelihood with a set of pre-collected trajectories. We then utilize MADDPG to train the searching agents to track the opponent agent. }
  \label{fig:system_flowchart}}
  \vspace*{-0.17in}
\end{figure*}
\noindent \textbf{Contributions:} In summary, our contributions are three-fold:
\begin{enumerate}
    \item We design a concise PMC filter structure that combines an \textit{a priori} model with a linear forward motion model, demonstrating its effectiveness in localizing targets.
    
    \item We propose a data-driven framework, utilizing the PMC filter in a MARL framework for a heterogeneous \ST{} team to track an intelligent evasive target in low observable settings (Figure \ref{fig:system_flowchart}).

    \item We showcase the effectiveness of our approach in tracking adversarial targets compared to baseline filters, expert policies, and other MADDPG baselines. Our method achieves an increase of 24.7\%, 46\%, and 51\% in localization accuracy, detection rate, and tracking performance, respectively.
\end{enumerate}

%% file: lit_review.tex
\section{Related Work \label{sec:RW}}
\subsection{Object Search and Track \label{sec:OST}}
Target search and tracking (\ST) has emerged as a prominent research topic in recent years \cite{multiaMultio, realConst, multiAUV, spiralLawn, interception, maddpgSearch, activeSearch, targePath, taxi, strategies}. 
Researchers have approached this problem by creating heuristic search strategies for agents such as spirals, lawn-mowers \cite{spiralLawn}, or interceptions \cite{interception}. However, these methods either assume knowledge of the target's exact location \cite{interception}, rely on a wide field of view \cite{activeSearch}, or are designed for non-evading, non-adversarial targets \cite{maddpgSearch, strategies}. 
Other work that assumes dynamic targets utilizes an exploration-exploitation dynamics (EED) strategy \cite{kwa2020optimal} based upon a particle swarm optimization algorithm \cite{kennedy1995particle} or information theoretic objective functions \cite{yang2023learning, yang2023policy}. In this paper, we demonstrate that our proposed learning-based approach surpasses both handcrafted heuristics and swarm searching strategies.



\subsection{Opponent Modeling}
Opponent modeling usually uses neural networks for target or opponent prediction. For instance,  DRON is proposed to model the behavior of opponents from opponent observations \cite{he2016opponent}. Grover et al. use imitation learning and contrastive learning to predict the next actions of an opponent \cite{grover2018learning}. Yu et al. utilize a model-based approach with neural networks for predicting the next state of the opponents \cite{yu2021model}. However, such approaches assume full observability of the opponent. 

In this paper, we propose a neural network filtering module (PMC) that estimates the current opponent location from limited detection and knowledge generalized from its previous behavior, to help the downstream MARL track an evasive target. Although traditional non-learning-based filters \cite{IMM0, IMM1, IMM2, dames2020distributed}
have been widely used for agent localization, such classic filters can neither incorporate prior knowledge nor generalize well for under-determined observation conditioning in long term settings since they usually require the difference between consecutive predictions \cite{intKalman, IMM0, taxi}. 

 

\subsection{Multi-Agent Reinforcement Learning (MARL) \label{sec:MARL}} 
MARL algorithms are designed to train multiple agents jointly to solve tasks in a shared environment. The simplest form is to train each agent independently with traditional RL algorithms (e.g. DQN\cite{dqn}, DDPG\cite{ddpg}). However, such an approach for training agent policies is not guaranteed to converge in a non-stationary multi-agent system. 

Rather than training each agent independently, researchers have utilized centralized training decentralized execution (CTDE) algorithms where all agent observations are used to learn a centralized critic, but each agent uses its own observations during execution \cite{maddpg, qmix}. Other solutions include utilizing learnable online communication channels such that agents can learn whom to communicate with, how to process messages, and when to communicate \cite{magic, esi_hetnet}. We build upon MADDPG \cite{maddpg} in our approach to train S\&T agents in an adversarial domain.

Utilizing MARL in adversarial environments is difficult as the policy of the opponent may evolve during training which leads to a distribution shift in the optimal policy for the search agents. AlphaStar \cite{vinyals2019grandmaster} partially solves the issue by training a ``league" of agents corresponding to different strategies such that each individual policy can focus on optimizing a certain aspect of the strategy space. Other approaches have used opponent modeling \cite{he2016opponent, grover2018learning, yu2021model, shiguang_multi_agent} to predict the future location of adversarial agents as additional loss terms during training. However, these approaches typically rely on continuous knowledge of opponent observations as input, which is not applicable in our setting. Our learnable PMC filter only conditions on the previous sparse detections and improves tracking performance on a continuous action, large state space, and partially observable domain compared to previous discrete action, small state space, and fully observable domains commonly used for MARL.



%% file: background.tex
\newcommand{\roboF}{g}
\newcommand{\roboBV}{\xi_u}
\newcommand{\camF}{h}
\newcommand{\camBV}{\zeta_u}
\newcommand{\camProj}{H}
\newcommand{\camPt}{r}

\newcommand{\imJac}{{L}}
\newcommand{\imJacV}{\boldsymbol{L}}

\newcommand{\geoJac}{{G}}

\newcommand{\numFeat}{n_{\text{F}}}

\newcommand{\pose}{\boldsymbol{P}}
\newcommand{\control}{\boldsymbol{u}}
\newcommand{\transCR}{^C\boldsymbol{T}_{R}}
\newcommand{\transCRv}{^C\boldsymbol{T}_{R,v}}
\newcommand{\transCRw}{^C\boldsymbol{T}_{R,\omega}}

\newcommand{\pointSet}{\boldsymbol{Q}}
\newcommand{\pointSetV}{\boldsymbol{q}}

\newcommand{\featureSet}{\boldsymbol{S}}
\newcommand{\featSetV}{\boldsymbol{s}}
\newcommand{\featSetDV}{\boldsymbol{s}^*}
\newcommand{\featSetDVdot}{\dot{\boldsymbol{s}}^*}
\newcommand{\featureSetA}{\boldsymbol{S}_a}
\newcommand{\featureSetD}{\boldsymbol{S}^*}
\newcommand{\featureSetDA}{\boldsymbol{S}_a^*}

\newcommand{\error}{\boldsymbol{e}}
\newcommand{\errorSet}{\boldsymbol{E}}

\newcommand{\imageJacobian}{\boldsymbol{L}}

\newcommand{\imageJacobiani}{\boldsymbol{L}_{S_i}}
\newcommand{\imageJacobianv}{\boldsymbol{L}_{S,v}}
\newcommand{\imageJacobianw}{\boldsymbol{L}_{S,\omega}}
\newcommand{\transFeature}{g}
\newcommand{\homo}{H}

\section{Environment}
\subsection{Partially Observable Markov Game}
A Markov Game is the multi-agent version of a Markov Decision-Process (MDP). In this work, we model adversarial search and track as a Partially Observable Markov Game (POMG)\MRSEdit{\cite{hansen2004dynamic}}. A POMG for N-agents can be defined by a set of states $\mathcal{S}$, a set of private observations for each agent \{$ \mathcal{O}_1, \mathcal{O}_2, \ldots, \mathcal{O}_N$\}, a set of actions for each agent \{$ \mathcal{A}_1, \mathcal{A}_2, \ldots \mathcal{A}_N$\}, a transition function $\mathcal{T} : \mathcal{S} \times \mathcal{A}_1 \times \ldots \times \mathcal{A}_N \mapsto \mathcal{S}$. At each timestep $t$, every agent \MRSEdit{indexed} $i, i \in \{1, \ldots, N\}$ receives an observation $O_i^t$, chooses an action $\MRSEdit{A_i^t} \in \mathcal{A}_i$, and obtains a reward $r_i^t: \mathcal{S} \times \mathcal{A}_i \mapsto \mathbb{R}$. The initial world state is drawn from a prior distribution $\rho$. In our environment (see \S\ref{sec:environment}), the RL policy has to explore a large search space with nearly no detection at the start, and the observation distribution is approximately static. However, the distribution will change when the policy becomes better and the detections increase. We expect our designed filter can help RL for an easier transition.

\subsection{Fugitive Escape Scenario}
\label{sec:environment}

In our evaluation of the proposed algorithm for adversarial \ST{}, we use a large-scale ($2428\times2428$, \MRSEdit{representing 51 x 51 km}) pursuit-evasion domain named Fugitive Escape scenario \cite{prev_work}\MRSEdit{\footnote{\MRSEdit{https://github.com/CORE-Robotics-Lab/Opponent-Modeling}}}. An intelligent evader \MRSEdit{with max speed 15 per timestep} is being tracked by a team of search agents in a partially observable environment. The \MRSEdit{game-ending} goal of the evader is to reach any hideout on the map, of which the pursuit team only knows a partial set. The terrain has varying visibility levels, with dense forest regions hindering the agents' detection abilities. Using $A^*$ search, the evader covertly makes their way to a hideout while traversing the shortest distance and taking advantage of dense cover. Additionally, the evader employs dynamic planning by using evasive behaviors to escape from the search team's contact.

The heterogeneous pursuit team consists of static (cameras) and dynamic agents (search parties and helicopters \MRSEdit{with max speed 20 and 127 respectively}) that collaborate to search and track the evader. Both the evading agent and the pursuit agents can only detect each other within a certain range, determined by their speeds and the terrain. 

\subsection{Heuristic Policy}
\label{sec:heuristic}
We create a heuristic policy for tracking the evader inspired by previous object searching works \cite{spiralLawn, interception}. The search agents can choose to either 1) go to the last known location of the evader, 2) intercept the evader, or 3) search in the vicinity of the evader's last known location. 
\begin{figure}
     \centering
        \subfloat[Intercept]{\includegraphics[width=0.15\textwidth]{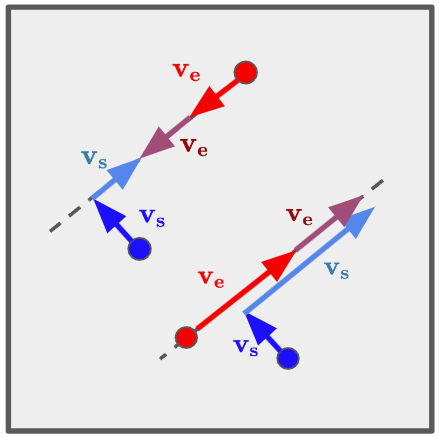}\label{fig:intercept}}
        \subfloat[Spiral]{\includegraphics[width=0.15\textwidth]{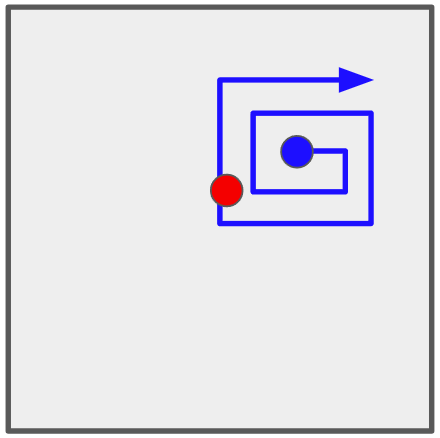}\label{fig:Spiral}}
        \vfill
        \vspace{-0.6em}
        \subfloat[Pointwise]{\includegraphics[width=0.15\textwidth]{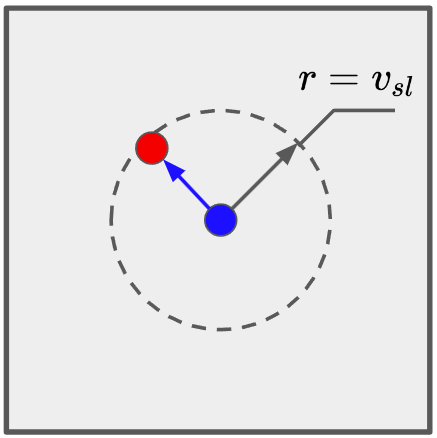}\label{fig:Pointwise}}
        \subfloat[Random Spiral]{\includegraphics[width=0.15\textwidth]{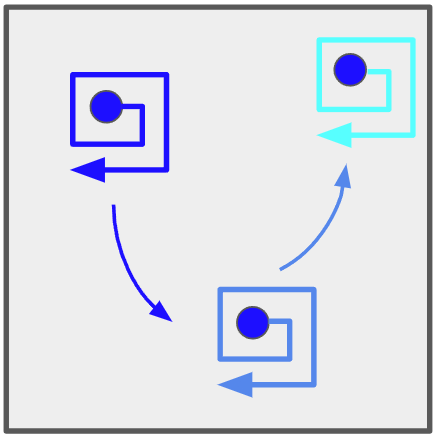}\label{fig:random_spiral}}
        \caption{Subpolicies used in the search agent heuristics.}
        \label{fig:heu_controllers}
        \vspace{-0.2in}
\end{figure}


In the Fugitive Escape domain, we have two types of dynamic agents--- search parties and helicopters, with speed limits of $v_{sl}$ and $v_{hl}$ respectively. The detection history of the evader is shared across all search agents and is denoted as $\lbrace d_i \rbrace _{i=1}^{N_{d}}$, where $d_i$ is the detected location of the evader and $N_{d}$ is the number of detections from the start of the episode. We denote the current detected location of the evader as $x_p$ and that of any search agent as $x_s$. 

If the evader is detected nearby and within reaching distance, (i.e. $\Vert x_p-x_s\Vert_{2}<v_{sl}$), the search agent will directly go to the detected location with the pointwise policy (Figure~\ref{fig:Pointwise}). Otherwise, the search agent will intercept by going to the interception point perpendicular to the opponent agent velocity (Figure~\ref{fig:intercept}) and then move along the vector towards the opponent's estimated location. If the search agent does not detect the evader, it will spiral around the estimated evader location for a set time (Figure~\ref{fig:Spiral}). If the agent still fails to find the evader, then the policy is reset and the search agent starts a spiral search at a random location (Figure~\ref{fig:random_spiral}).

%% file: method.tex
\newcommand{\mat}{\boldsymbol}

\section{Method}\label{sec:method}


 We propose a novel approach named Prior Motion Combined MADDPG (PMC-MADDPG) to improve the efficiency of solving \ST{} problems in sparse detection environments (Figure  \ref{fig:system_flowchart}). PMC-MADDPG is a framework that embeds a filtering model (PMC), which estimates the current location of the adversary, into MADDPG. We show that the filtering model aids in improving MARL performance even with sparse information at the start of training (\S\ref{sec:exp}). We take inspiration from classical filters, where most classical methods merge different information sources (e.g. prediction and update, multiple motion models, etc.) with weights (e.g. Kalman gain) for some optimal loss (minimum mean-square error). Following this pattern, we build our filter as two different information branches (a prior network and a linear motion model) and a learnable weight $\alpha$.


\subsection{PMC Filter System Design}\label{sec:filter_theory}
Our proposed Prior Motion Combined (PMC) filtering model (Figure \ref{fig:system_flowchart}) consists of three parts 1) a learned prior network to inform the search agents of the adversarial agent preferences, 2) a linear motion model predictor to provide estimates of the adversarial motion and 3) a confidence network that learns to weigh these components relative to the input observation. 

\subsubsection{Prior Network} 

The adversary may exhibit distinct patterns or preferences while meeting its objectives such as evading surveillance agents, optimizing travel costs or \MRSEdit{heading to} different hideouts.
We utilize a learning based approach to encode data from past trajectories of the adversary \MRSEdit{into an embedding space}. 
Fully connected \MRSEdit{(FC)} network is used as an encoder that takes as input the starting location of the adversary and the current timestep from the observation:
\begin{equation}
\boldsymbol{p_h} = Encoder(\boldsymbol{x_1, t}) \label{eq:encoder} \\
\end{equation}
where $\mat{p_h}$ is the embedding from the encoder, $\boldsymbol{x_1}$ is the starting location of the adversary and $t$ is the current timestep.

\subsubsection{Motion Model}

We use a linear motion model where the input is the timestep and the last two detected locations. We output the estimated current location based \MRSEdit{on} the constant velocity assumption. The state of the adversarial agent at timestep $k$ can be estimated as:
\begin{equation}\label{eq:kalman_adv}
\mat{x}_k = \mat{x}_{\hat{k}} + (k - \hat{k})\mat{u}_{\hat{k}}
\end{equation}
where $\mat{x}_k$ and $\mat{u}_k$ are the state (location) and control input (velocity); $\hat{k}$ is the latest timestep where detection occurred. This can be shown to satisfy the intermittent Kalman filter estimation \cite{intKalman}. We pass the output of the motion model $\mat{x_k}$ through several fully connected layers to produce an embedding $\mat{m_h}$ for the motion prediction to be used for the decoder described in the following subsection \S\ref{subsub:conf}. 

\subsubsection{Confidence Network}\label{subsub:conf}

We use an \MRSEdit{FC} {\textit{confidence net}} to generate \MRSEdit{a scalar} weight ${\alpha}$ from the observation to balance the prior and motion outputs:
\begin{equation}
\mat{G}_e = {\alpha} \mat{p}_h + ({1} - {\alpha}) \mat{m}_h
\end{equation} 
where $\mat{p}_h$, $\mat{m}_h$ and $\mat{G}_e$ are prior, motion and Gaussian embeddings respectively and ${\alpha}$ is the learnable weight.

Since the adversary's path preferences are encoded in the prior network, the decoder can produce multiple hypotheses (mixture of Gaussians) for state estimation:
\begin{equation}
\MRSEdit{\mat{\lambda}_i}, \mat{\mu}_i, \mat{\sigma}_i = Decoder(\mat{G}_e), i=1, 2, 3, ..., N_g
\end{equation}
where \MRSEdit{$N_g=8$} is the number of components in the Gaussian mixture and \MRSEdit{$\mat{\lambda}_i$}, $\mat{\mu}_i$ and $\mat{\sigma}_i$ are the mixture weight, mean and variance of the mixture component $i$. We utilize the negative log-likelihood loss (Equation~\ref{eq:lll}) for a mixture of Gaussians to train the entire network including the prior network, layers encoding the motion model, and decoder: 
\begin{align}
\label{eq:lll}
    \MRSEdit{\mathcal{L} = -\text{log}(\sum\nolimits_{i=1}\nolimits^{N_g}\mat{\lambda_i}\cdot p(\mat{y}|\mat{\mu_i}, \mat{\sigma_i}^2))}
\end{align}
\MRSEdit{where $\mat{y}$ is the true adversary location normalized to $[0,1]$.} 
\begin{figure*}[t!]
	\centering
	\includegraphics[scale=0.35]{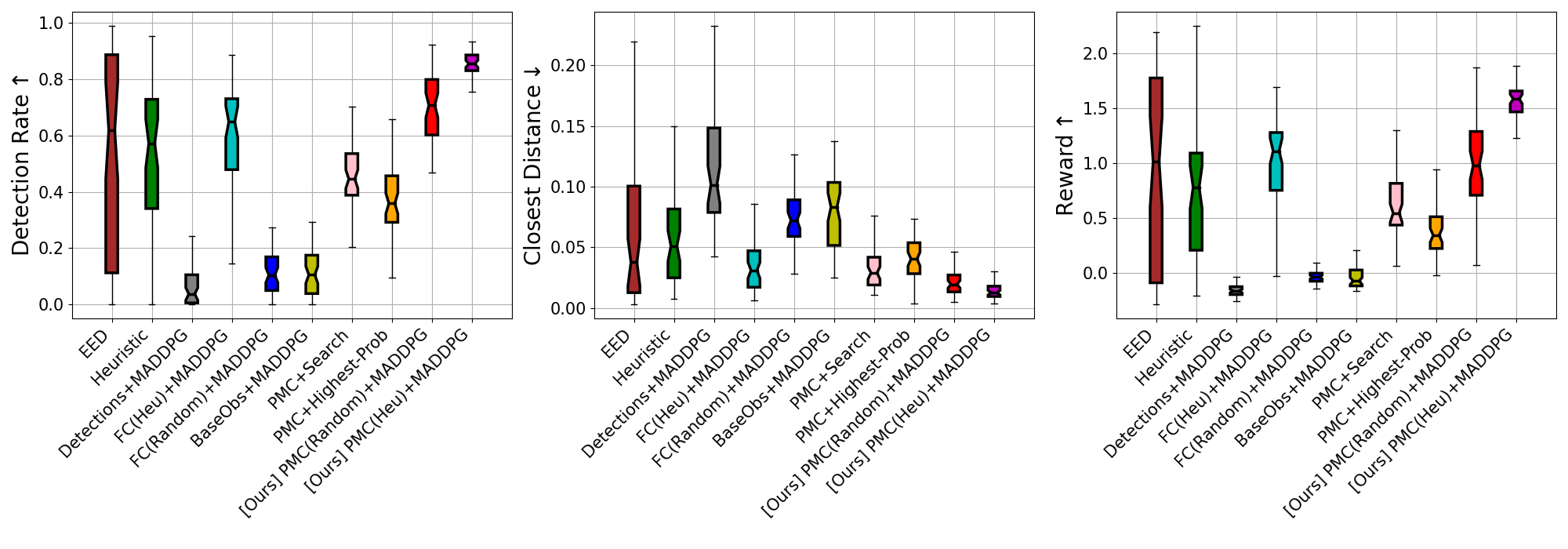}
  \vspace*{-0.38in}
	\caption{\toedit{MARL + Filtering evaluation results: we show our PMC filtering model + MADDPG outperforms all other baselines. Higher detection rate and reward, and lower closest distance means better performance.}}
	\label{fig:marl_metrics}
   \vspace*{-0.22in}
\end{figure*}

\color{black}
  
\tikzset{global scale/.style={
    scale=#1,
    every node/.append style={scale=#1}
  }
}

\subsection{PMC Filter Informed MARL}\label{sec:MADDPG}

In this section, we discuss how the PMC Filter described in Section \ref{sec:filter_theory} is utilized in the MARL formulation to improve search and tracking. We hypothesize that our RL based method (PMC-MADDPG) can perform better than hand crafted heuristics \cite{spiralLawn, interception} because the policy is capable of adapting to changing environment state distributions.

In traditional MADDPG, the value function faces the challenge of inferring the evasive agent's location from observations and rewards in a noisy and sparsely observable environment. This inference becomes particularly difficult due to the limited exploration of the state space. However, our PMC filter, trained directly with a supervised loss using historical trajectories, offers a solution. By augmenting the input to the actor-critic with an estimate of the evading agent's location, the PMC model helps alleviate the burden of this challenging inference. As a result, we propose that the PMC model effectively informs MARL training and enhances both sample efficiency and exploration capabilities.





\subsubsection{Multi-agent deep deterministic policy gradient (MADDPG)}\label{sec:MADDPG_eqations}
MADDPG is a multi-agent extension of the basic Deep Deterministic Policy Gradient (DDPG) reinforcement learning algorithm and follows the off-policy centralized training and decentralized execution paradigm \cite{maddpg}. Each agent has its own actor-critic structure but the critic takes the combined state-action pairs of all agents compared with the input of the actor which is each its own experience tuple. 
 
We use MADDPG as our framework to train the search agent policies. For each search agent, the base observation $o_b$ includes its own current location, the current timestep, the current locations of all other searching agents, and the location of known hideouts. The augmented observation $o_a$ is the extra information appended to each agent's observation. The PMC-MADDPG uses the estimated opponent location from the filter relative to the search agent as $o_a$
(Equation \eqref{eq:gaussians}):
\begin{equation}\label{eq:gaussians}
\mu^{loc}_{j} = \mu_j - x_s, j=1,2,3,...,N_g
\end{equation} 
Each agent $i$ receives its own observation denoted as $o_i = [o_b \Vert o_a]$. Based on this observation, each agent independently determines its action using its own \MRSEdit{FC deterministic} policy:
\begin{equation}
a_i = \pi_i(o_i), i=1,2,...,N_a
\end{equation}
where $N_a$ is the number of search agents.
The critic is then updated to minimize the TD error ($L(\theta_{i})$) between two consecutive timesteps:
 \begin{equation}
 \begin{split}
 L(\theta_{i})=\mathbb{E}_{\MRSEdit{O},a,r,O^{\prime}}[(Q_{i}^{\pi}(\MRSEdit{O},a_{1},...,a_{N_a})-y)^{2}]
 \end{split}
 \end{equation}
 \begin{equation}
 \begin{split}
 y=r_{i}+\gamma Q_{i}^{\pi^{\prime}}(O^{\prime},a_{1}^{\prime},...,a_{N_a}^{\prime})\MRSEdit{\vert_{a_{i}^{\prime}=\pi_{i}^{\prime}(o_{i})}} \label{eq:critic_update}
 \end{split}
 \end{equation}
Then the policy is updated with the gradient:
 \begin{multline}
 \nabla_{\MRSEdit{\xi_{i}}}J(\xi_{i})
 =\mathbb{E}_{O,a\sim \mathcal{D}}[\nabla_{\xi_{i}}\pi_{i}(a_{i}\vert o_{i}) \\ 
 \nabla_{a_{i}}Q_{i}^{\pi}(O,a_{1},...,a_{N})\MRSEdit{\vert_{a_{i}=\pi_{i}(o_{i})}}] \label{eq:actor_update}
 \end{multline}
 where $\gamma$ is discount factor. $\theta_{i},\xi_{i}, i=1,2,...,N_a$ are the parameters of the  critic and actor neural networks $Q_{i}, \pi_{i}$ respectively. $O$ is the concatenation of all agents' observation. We assume the action $a_i$ to be the search agent velocity. Samples are randomly drawn from the replay buffer $\mathcal{D}$ to train the actor-critic and all the gradients are cut off at the observation, which means we only update the network parameters using equations \eqref{eq:critic_update} and \eqref{eq:actor_update} while the PMC is updated only with \eqref{eq:lll}. The superscript prime means the quantity is of the next timestep and the subscript indicates the agent index. We train the algorithm with \MRSEdit{each agent's} detection reward and a distance shaping term as \cite{maddpg}.
\newcommand{\trajTimeStart}[1]{t_{#1,s}}
\newcommand{\trajTimeEnd}[1]{t_{#1,e}}
\newcommand{\featureTraj}[1]{\featureSetD_{#1} (t) |_{\trajTimeStart{#1}}^{\trajTimeEnd{#1}}}

\tikzset{global scale/.style={
    scale=#1,
    every node/.append style={scale=#1}
  }
}

\subsubsection{PMC Filter in MADDPG}

\toedit{The evading agent is rarely detected at the start of the MARL because the search agent policies are not trained well. Therefore we hypothesize that our PMC filter can compensate for the information loss by providing its position estimation to foster training. Specifically, we concatenate the Gaussian parameters $O_a = \{\MRSEdit{\lambda_i}, \mu_i, \sigma_i\}_{i=1}^{N_g}$ from the filter to the end of the base observation $O_b$ as described in \S\ref{sec:MADDPG_eqations} and compare the results with other augmented observations. }


We evaluate the benefit of augmenting MADDPG with a filter module by comparing the performance on two datasets -- random and heuristic depending on how the search agents interact with the evader when pretraining the filter. In the \textbf{random} dataset, the search agents use random policies such that there are nearly no detections and the evader does not need to actively escape. In the \textbf{heuristic} dataset, the search agents use heuristics such that the evader has interactions with the searching team. 
We show that our approach outperforms all baselines in both datasets in \S\ref{sec:exp}.


%% file: result.tex
\section{Experiments and Results}\label{sec:exp}

This section shows our proposed PMC filter model can successfully estimate the evader location and aid in downstream MADDPG training. We divide our comparisons into three groups: 1) PMC filter vs. RNN-based baseline filters, 2) MARL-based method vs. heuristics, and 3) validation of the PMC filter's positive contribution to MARL. We report the MARL performance averaged over 50 trajectories and show all experiment results in Table~\ref{tb:pmcMaddpgRes}, Figures~\ref{fig:marl_metrics}-\ref{fig:total_reward} where closest distance indicates the distance between the evader and the closest search agent ($\downarrow$ means lower is better). We denote the search agent policies used to warm-start the filter as either a heuristic policy (Heu) or random policy (Random). 

\subsection{PMC Filter Validation}\label{subsec:pmc_filter}
We first assess the accuracy of our PMC filter by comparing it to several baseline filters on prior datasets independent of MARL. Our PMC filter outperforms four RNN-based filters and one FC configuration (Table \ref{tb:randomBlueFilter}-\ref{tb:heuBlueFilter}) on various metrics. RT refers to the forward running time for one batch. We evaluate the log-likelihood (LL) of the adversary location given the predicted Gaussian distribution. ADE is the $l_2$ distance between the mean of the predicted distribution and the adversary location. Confidence Threshold Percentage (CTP) is the ratio of timesteps where the model predicts a probability of the adversary location greater than $0.5$.
\begin{table}[ht]
  \vspace*{-0.09in}
  \addtolength{\tabcolsep}{-2.5pt}
  \centering
  \begin{minipage}[t]{0.49\textwidth}
    \centering
    \caption{\textrm{Random Search Policy Evaluation Metrics (Mean)}%
    \vspace*{-0.1in}
    \label{tb:randomBlueFilter}}
    \begin{tabular}{|c||cccc|}
    \hline 
    \textbf{Filt.} & RT$\downarrow$ & LL$\uparrow$ & ADE$\downarrow$ & CTP$\uparrow$\\ 
    \hline 
    LSTM  & 8.93e-4  & 1.74  & 1.38e-1  & 4.27e-1\\ 
    GNN & 7.14e-3  & 2.93  & 7.91e-2  & 5.12e-1\\ 
    VRNN & 8.85e-3  & 4.70  & 4.59e-2  & 9.01e-1\\ 
    VRNN-GNN & 1.51e-2  & 3.61  & 6.13e-2  & 7.14e-1\\ 
    FC & \textbf{5.02e-4}  & 5.73  & 5.74e-2  & 9.63e-1\\
    PMC [ours]  & 9.17e-4  & \textbf{7.25}  & \textbf{4.19e-2}  & \textbf{9.71e-1}\\
    \hline 
    \end{tabular}
    \vspace{1em}
  \end{minipage}
  \begin{minipage}[t]{0.49\textwidth}
    \centering
    \caption{\textrm{Heuristic Search Policy Evaluation Metrics (Mean)}%
    \vspace*{-0.1in}
    \label{tb:heuBlueFilter}}
    \begin{tabular}{|c||cccc|}
    \hline 
    \textbf{Filt.} & RT$\downarrow$ & LL$\uparrow$ & ADE$\downarrow$ & CTP$\uparrow$\\ 
    \hline 
    LSTM  & 8.88e-4  & 1.74  & 1.28e-1  & 4.43e-1\\ 
    GNN & 7.56e-3  & 4.74  & 5.23e-2  & 8.11e-1\\ 
    VRNN & 9.11e-3  & 5.13  & 5.46e-2  & 8.67e-1\\ 
    VRNN-GNN & 1.42e-2  & 5.33  & 7.69e-2  & 8.07e-1\\ 
    FC & \textbf{4.90e-4}  & 6.15  & 4.16e-2  & 8.92e-1\\
    PMC [ours] & 9.24e-4  & \textbf{7.57}  & \textbf{3.41e-2}  & \textbf{9.66e-1}\\
    \hline 
    \end{tabular}
  \end{minipage}

  \addtolength{\tabcolsep}{2pt}
  \vspace*{-0.09in}
\end{table}

We evaluate all models on two datasets: one with random search agent policies (Table \ref{tb:randomBlueFilter}) and one with heuristic search agent policies (Table \ref{tb:heuBlueFilter}). All filter configurations follow an encoder-decoder structure where the encoder is indicated in Table \ref{tb:randomBlueFilter}-\ref{tb:heuBlueFilter} and the decoder is an FC layer that outputs the Gaussian mixture parameters. Both PMC and FC models use the same input (the last two detections with velocities, current timestep, and starting location of the evader). The input for all RNN-based filtering approaches additionally includes the observation histories of all agents.  Results clearly show that only the FC model is comparable with our PMC model but is still worse by 19.84\% and 30.26\% on LL and ADE respectively, averaged across both datasets. Next, we show the results by incorporating the FC and PMC models with our MARL framework for adversarial \ST{}.
\subsection{Advantages of MARL over Heuristic Policies}
We use two non-learning-based approaches as baselines--- 1) Heuristic, which combines commonly used search strategies like spirals, pointwise moving, and interceptions (see \S\ref{sec:heuristic}), and 2) Exploration Exploitation Dynamics (EED) which is a specialized dynamic target search and tracking \cite{kwa2020optimal} method. Both approaches have comparable results achieving greater than 50\% detection rate. Our results show the MADDPG-based approaches outperform the non-learning-based heuristics by at least 15\% and 36\% in terms of detection rate and closest distance (Table \ref{tb:pmcMaddpgRes}). We show that the learning-based MADDPG approaches are able to adapt better to the distribution of the adversarial environment compared to hand-crafted heuristics. 

As an ablation study, we construct two hand-crafted search agent policies that use the PMC filter but \textit{do not learn}. PMC(Heu) + Highest-Prob drives the search agents to the highest probability Gaussian while PMC(Heu) + Search splits the searching agents amongst multiple Gaussians. Our results show that the MADDPG policy can evolve as the distribution shifts (e.g., more evasive behavior when detected) compared to the deterministic heuristics even though they have same access to prior knowledge within the PMC filter.

\begin{table}
  \vspace*{0.06in}
  \addtolength{\tabcolsep}{-2.5pt}
\centering
  \begin{minipage}[t]{0.485\textwidth}
    \centering
   \caption{\textrm{Combination of Different Filters and Policies \\ (Mean $\pm$ Standard Deviation)}}
   \vspace*{-0.1in}
   \label{tb:pmcMaddpgRes}
  \begin{tabular}{|l||ccc|}
    \hline 
    {\textbf{Filter \& Policy}} &{\textbf{Detect. Rate}} & {\textbf{Closest Dist.}} & {\textbf{Reward}} \\ 
    \hline 
    \textbf{Non-Learning Approaches} & & & \\
    Heuristic & 0.52$\pm$0.28  & 0.06$\pm$0.05  & 0.82$\pm$0.73\\ 
    EED & 0.51$\pm$0.38  & 0.07$\pm$0.08  & 0.83$\pm$0.90\\ 
    PMC(Heu)+Highest-Prob & 0.38$\pm$0.16  & 0.04$\pm$0.02  & 0.44$\pm$0.40\\
    PMC(Heu)+Search & 0.47$\pm$0.13  & 0.03$\pm$0.01  & 0.67$\pm$0.38\\
    \textbf{Learning Approaches} & & & \\
    BaseObs+MADDPG & 0.11$\pm$0.09 & 0.08$\pm$0.03 & -0.04$\pm$0.09\\ 
    Detections+MADDPG & 0.06$\pm$0.07  & 0.12$\pm$0.06  & -0.16$\pm$0.05\\ 
    FC(Random)+MADDPG  & 0.11$\pm$0.08 & 0.08$\pm$0.03 & -0.04$\pm$0.05\\ 
    FC(Heu)+MADDPG & 0.60$\pm$0.19  & 0.04$\pm$0.03 & 1.01$\pm$0.42\\    
    \textbf{Our Approches} & & & \\
    PMC(Random)+MADDPG & 0.69$\pm$0.15  & 0.02$\pm$0.01  & 1.00$\pm$0.38\\
    PMC(Heu)+MADDPG & \textbf{0.84}$\pm$0.12  & \textbf{0.02}$\pm$0.04  & \textbf{1.53}$\pm$0.27\\
    \hline 
  \end{tabular}
  \vspace*{-0.1in}
  \end{minipage}
\end{table}

\subsection{Model Based Filtering Benefits}\label{sec:betterThanOtherAug}
We compare the results of utilizing a PMC model and those without a filter model. The BaseObs+MADDPG uses the null vector $\emptyset$ while Detections+MADDPG uses the last two detections (with velocities) $o_d = \{x_i, v_i\}_{i=1}^{2}$ as the augmented observation $o_a$. We show that our PMC models outperform these baselines, performing on average 88\% better on detection rate and achieving 41 times as much reward. While the input to the agents is the same, the filtering models provide the location estimates from an extra network that the baselines do not have. We confirm that providing tabula rasa agents with an opponent model from detection history enables them to learn within our large, partially observable domain. 

\begin{figure}[h]
\vspace*{-0.1in}
	\centering
	\includegraphics[scale=0.33]{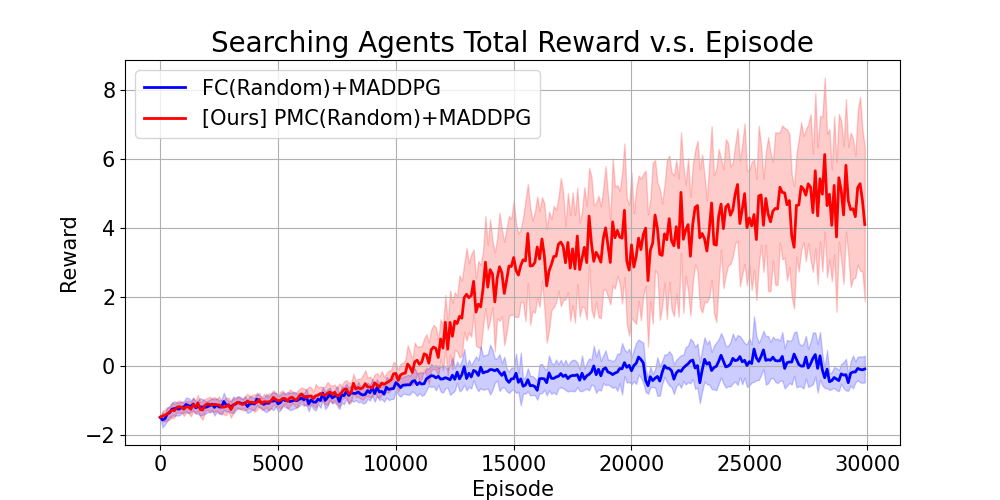}
        \includegraphics[scale=0.33]{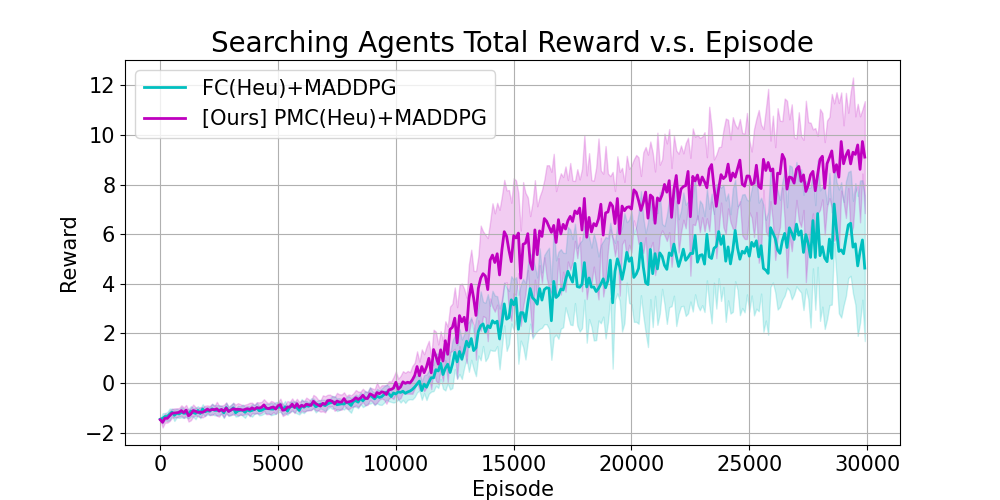}
	\caption{We sum all searching agents rewards and plot the total reward curve versus the training episodes. The PMC based MADDPG gains reward faster and higher. The filters trained on heuristic dataset perform better than those trained on random dataset.}
 \vspace*{-0.17in}
	\label{fig:total_reward}
\end{figure}
\subsection{Comparing PMC with FC Filters within MADDPG}
In this section, we analyze how the PMC filtering model performs against the FC when used within MADDPG. We use FC and PMC augmented MADDPG, which are the best filters from Section \ref{subsec:pmc_filter}. We find that the FC+MADDPG models underperform our PMC+MADDPG models by 53.59\% on detection rate. Thus, the improvement in the MADDPG performance is due to the PMC filter structure since both PMC and FC filters have the same inputs. Reward curves in Figure \ref{fig:total_reward} show that the PMC+MADDPG achieves better sample efficiency and higher reward compared with FC+MADDPG. Further, the PMC filter trained on the heuristic dataset (PMC(Heu)) can achieve higher performance since the filtering module also encodes the interactions between search and evader agents.
\begin{figure}[h]
\captionsetup[subfigure]{labelformat=empty}
     \centering
        \rotatebox{90}{\parbox[c][0.5cm][c]{2.4cm}{\small \centering Heuristic}}
        \subfloat[]{\includegraphics[width=0.14\textwidth]{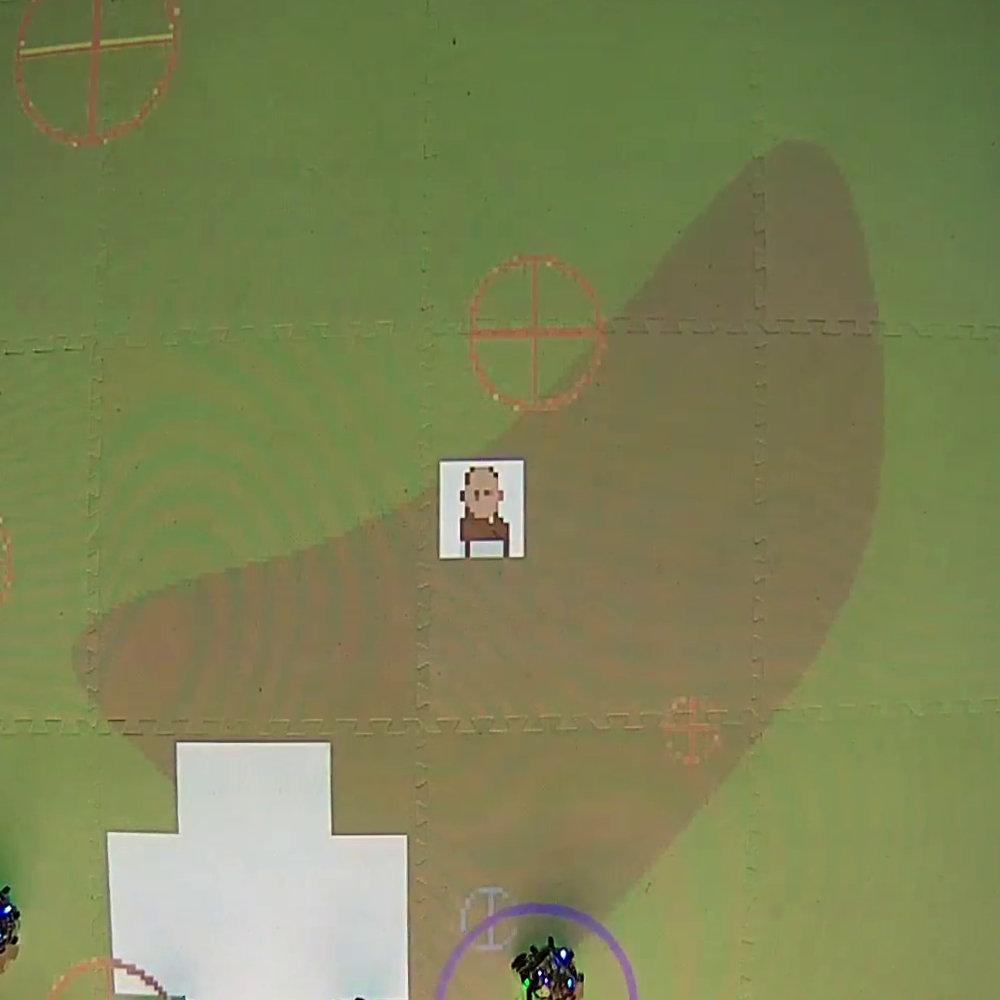}\label{fig:intecept}
        \includegraphics[width=0.14\textwidth]{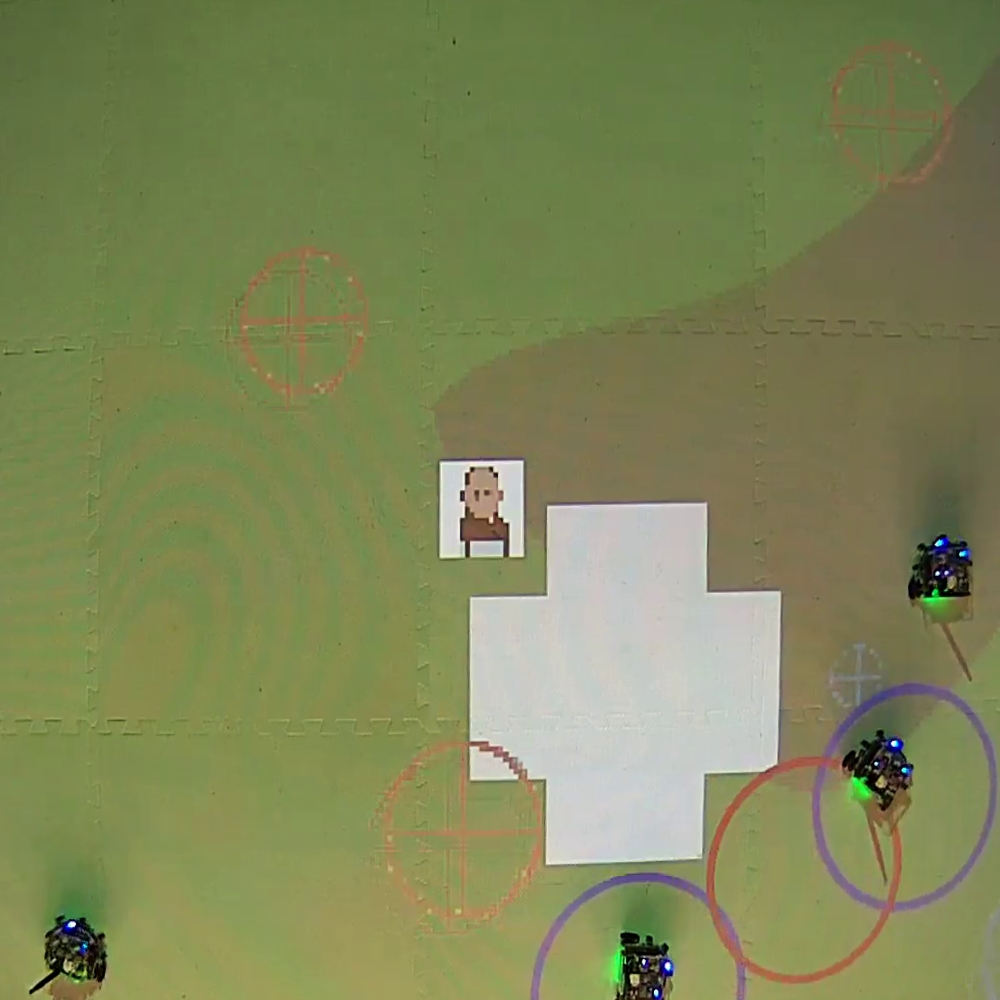}
        \includegraphics[width=0.14\textwidth]{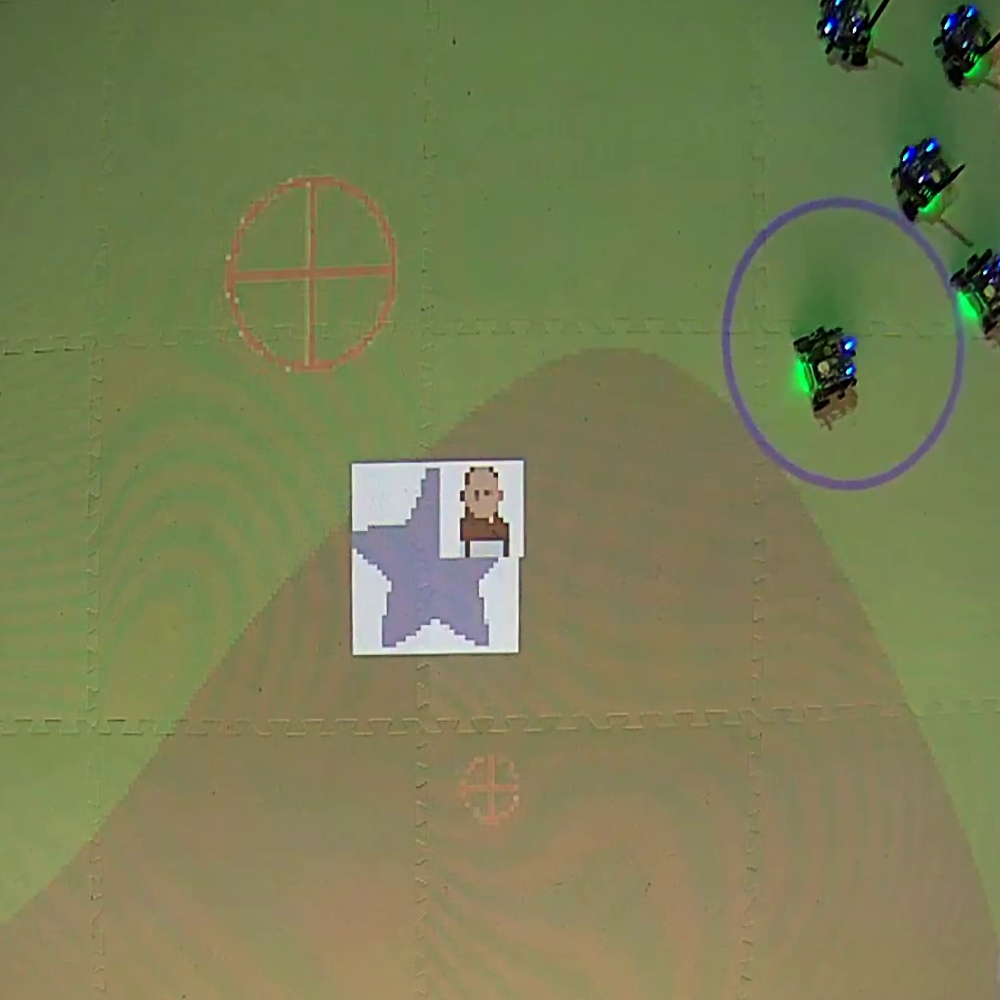}}
        \vfill
        \rotatebox{90}{\parbox[c][0.5cm][c]{2.4cm}{\small \centering PMC+MADDPG}}
        \subfloat[]{\includegraphics[width=0.14\textwidth]{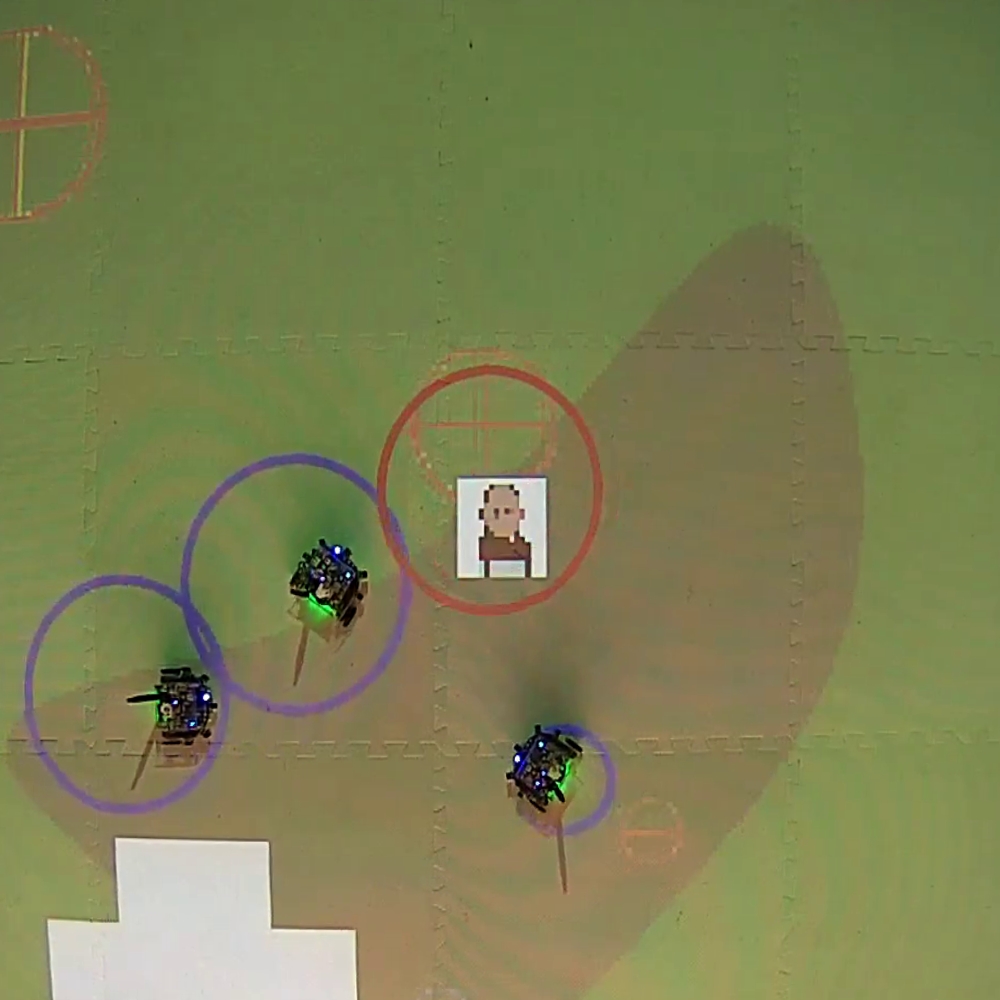}\label{fig:spiral}
        \includegraphics[width=0.14\textwidth]{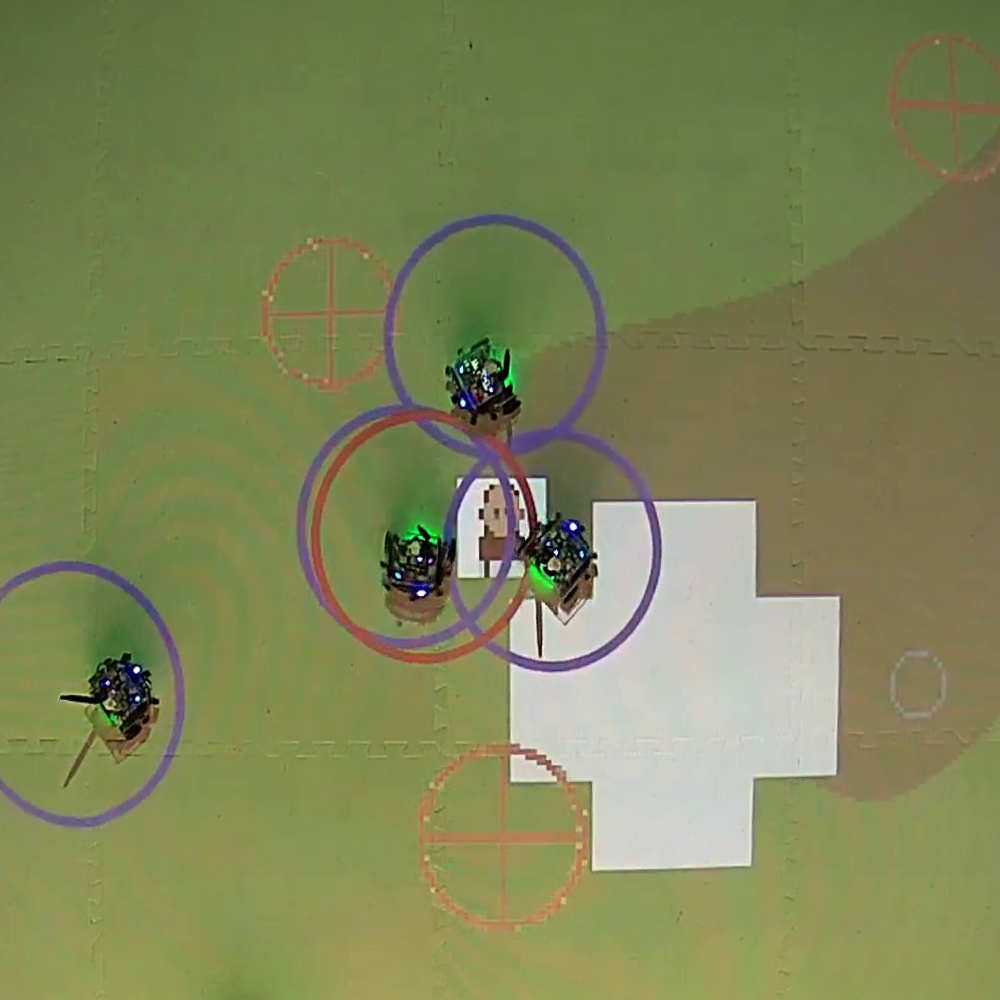}
        \includegraphics[width=0.14\textwidth]{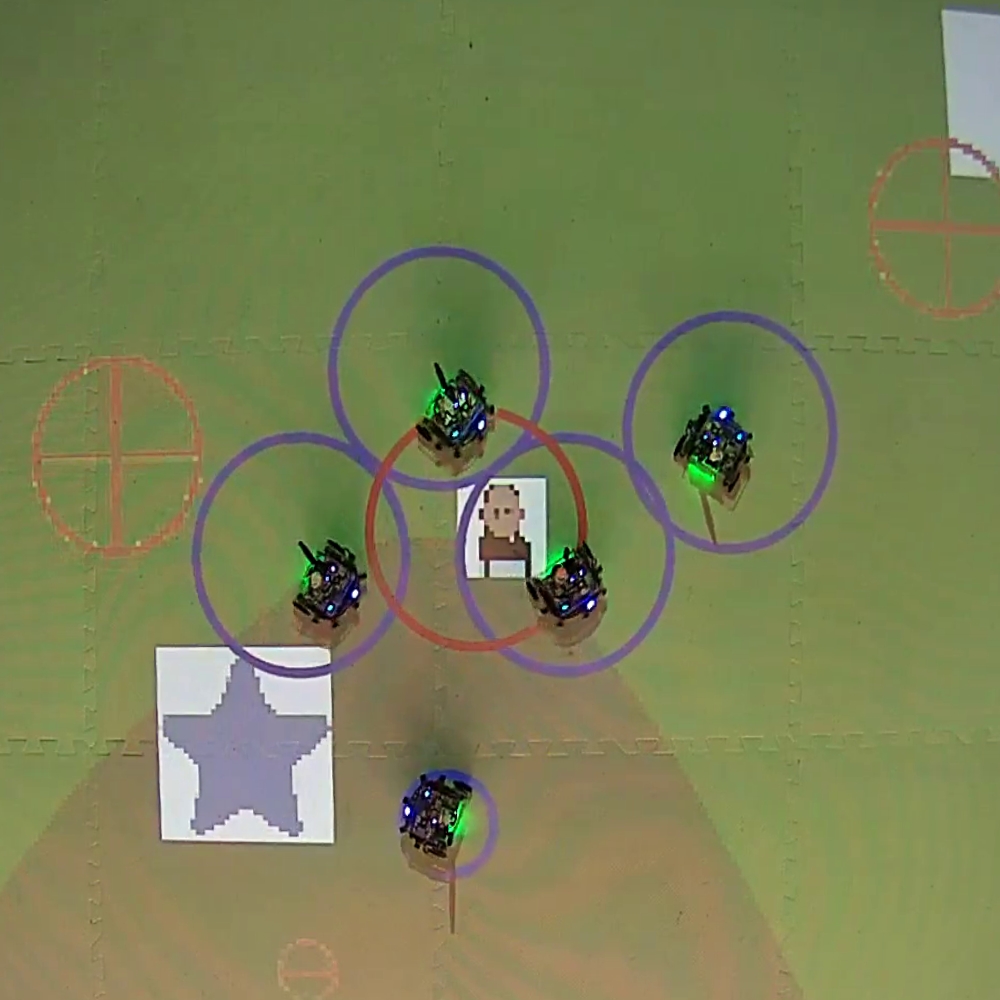}}
        \vfill
        \vspace{-0.05in}
        \parbox[c][0.0cm][c]{\linewidth}{\raggedleft $\boldsymbol{\xrightarrow[]{\hspace*{6.8cm}}}$ Time}
        \caption{Demonstration of Heuristic vs PMC+MADDPG(Heu) on real robots. The star and large white region represent the hideout and obstacle. Darker regions of the map represent dense forests where visibility is low.}
         \vspace*{-0.1in}
        \label{fig:robotarium}
\end{figure}
\subsection{Real Robot Demonstration}
We also conducted tests using the Robotarium \cite{8960572} testbed to evaluate the performance of our PMC+MADDPG method in real-world scenarios with dynamic and collision constraints. The results, as depicted in Figure \ref{fig:robotarium}, highlight the comparison between the heuristic policy and our method. In these experiments, the ground robots represent the ground search parties, with blue circles indicating their detection range. Due to the unavailability of drones, we projected the helicopter (represented by a red circle) onto the test bed. The findings clearly demonstrate that the search agents following the heuristic policy fail to capture the evader by erroneously navigating to the wrong side of obstacles. Conversely, our PMC model, which proposes multiple potential evader positions, effectively assists the MARL policy in continuous tracking until the evader reaches their hideout.

%% file: conc.tex
\section{Discussion and Conclusion}
We propose a novel approach for improving the efficiency and effectivenes of \ST{} systems in sparse detection environments by leveraging a joint framework that utilizes a filtering module (PMC) within MADDPG. While the proposed approach demonstrated promising results, there exist a few limitations that can be addressed in future work. First, our current approach is aimed at tracking a single adversary, and we aim to extend our approach to handle multiple adversaries. Second, our current approach requires previous interactions to warmstart the filtering module, which we intend to remove. \MRSEdit{We will also further try to apply the approach in drug tracing or disasters rescuing tasks.}